\documentclass[conference]{IEEEtran}
\IEEEoverridecommandlockouts
\usepackage{cite}
\usepackage{amsmath,amssymb,amsfonts}
\usepackage{algorithmic}
\usepackage{graphicx}
\usepackage{textcomp}
\usepackage{color,xcolor}
\usepackage{subfigure}
\usepackage{array}
\usepackage{booktabs}
\usepackage{url}
\usepackage{soul}
\usepackage{lineno}

\usepackage[ruled,vlined]{algorithm2e}
\SetCommentSty{mycommfont}
\SetKwInput{KwInput}{Input}                
\SetKwInput{KwOutput}{Output}              
\SetKwInput{KwPar}{Parameter}              
\usepackage{verbatim}
\usepackage{float}
\usepackage{amssymb}

\UseRawInputEncoding
\def\BibTeX{{\rm B\kern-.05em{\sc i\kern-.025em b}\kern-.08em
		T\kern-.1667em\lower.7ex\hbox{E}\kern-.125emX}}
\begin{document}
	
	\title{
		Attention-Driven LPLC2$\,$Neural Ensemble$\,$Model$\,$for Multi-Target Looming Detection and Localization
	}
	\author{
		\IEEEauthorblockN{Renyuan Liu$^{1}$, Qinbing Fu$^{*,1,2}$}
		\IEEEauthorblockA{$^{1}$ School of Mathematics and Information Science, Guangzhou University, Guangzhou, China}
		\IEEEauthorblockA{$^{2}$ Machine Life and Intelligence Research Center, Guangzhou University, Guangzhou, China}
	}
	
	\maketitle
	
	\begin{abstract}
		Lobula plate/lobula columnar, type 2 (LPLC2) visual projection neurons in the fly's visual system possess highly looming-selective properties, making them ideal for developing artificial collision detection systems. 
		The four dendritic branches of individual LPLC2 neurons, each tuned to specific directional motion, enhance the robustness of looming detection by utilizing radial motion opponency. 
		Existing models of LPLC2 neurons either concentrate on individual cells to detect centroid-focused expansion or utilize population-voting strategies to obtain global collision information. 
		However, their potential for addressing multi-target collision scenarios remains largely untapped. 
		In this study, we propose a numerical model for LPLC2 populations, leveraging a bottom-up attention mechanism driven by motion-sensitive neural pathways to generate attention fields (AFs). 
		This integration of AFs with highly nonlinear LPLC2 responses enables precise and continuous detection of multiple looming objects emanating from any region of the visual field. 
		We began by conducting comparative experiments to evaluate the proposed model against two related models, highlighting its unique characteristics. 
		Next, we tested its ability to detect multiple targets in dynamic natural scenarios. 
		Finally, we validated the model using real-world video data collected by aerial robots. 
		Experimental results demonstrate that the proposed model excels in detecting, distinguishing, and tracking multiple looming targets with remarkable speed and accuracy. 
		This advanced ability to detect and localize looming objects, especially in complex and dynamic environments, holds great promise for overcoming collision-detection challenges in mobile intelligent machines.
	\end{abstract}
	

	\section{Introduction}
	\label{Sec: introduction}
	
	The ability to quickly and accurately identify approaching objects is critical for the survival of animals \cite{1999_LGMD_Gabbiani}. 
	Various species accomplish this by distinguishing between different types of motion based on the unique movement characteristics of objects \cite{1962_monky_looming,2008_avoidance_crabs,2014_avoiding_flies}. 
	These efficient collision detection mechanisms also hold significant promise for applications in autonomous driving, drone technology, and robotic navigation. 
	However, machine learning-based collision detection methods often require large training datasets \cite{2022_shallow_learning_Zhou,2021_detection_deep}, and sensor-based approaches, such as LiDAR, are typically expensive and limited in application contexts \cite{2023_detection_UAV}.
	
	Insects exhibit simple yet highly effective collision avoidance systems \cite{2014_avoiding_flies}. 
	Inspired by the lobula giant movement detector (LGMD) neurons in locusts, researchers have developed low-power, fast, and real-time collision detection systems \cite{2005_robot_Yue,2016_Colias_Hu}. 
	However, traditional LGMD-based bio-inspired neural networks face limitations, particularly in their selectivity. 
	These networks struggle to precisely differentiate looming motion from other movement types, often responding to receding or translating movements \cite{2019_review_Fu}.
	
	Collision detection systems based on directionally sensitive neurons in fruit flies \emph{Drosophila} have shown promising results \cite{2021_multiple-region_fly,2024_pathway_Song}. 
	Recent biological studies suggest that the lobula plate/lobula columnar, type-2 (LPLC2) visual projection neurons in fruit flies are better suited for looming perception due to their ultra-selective properties \cite{2017_LPLC2_klapoetke,2022_LCs_klapoetke}. 
	These neurons, with cross-shaped dendrites spanning specific layers, respond strongly to outward motion from the center of their receptive field while being inhibited by inward motion \cite{2017_LPLC2_klapoetke}.
	
	Recent studies have investigated implementing LPLC2's extreme selectivity through nonlinear integration \cite{2022_LPLC2_Hua_Mu,2023_LPLC2_Wu} and machine learning techniques to predict their properties and tuning \cite{2022_shallow_learning_Zhou}. 
	While these approaches highlight the effectiveness of bio-inspired LPLC2 models for collision detection, they overlook a critical aspect: the LPLC2 neural ensemble consists of numerous neurons, each covering a portion of the visual field. 
	This raises two key questions: 
	(1) How can LPLC2 neurons collectively recognize expansion originating from any location within the visual field? 
	(2) How can the responses of the LPLC2 population be harnessed to achieve multi-target collision detection?
	
	In this study, we introduce a multi-attention mechanism to the LPLC2 neural network model (mLPLC2) to address these challenges. 
	Inspired by the LPLC2 population's ability to cover the entire visual field of the fruit fly (Fig. \ref{Figure_LPLC2_Physiology}), 
	we develop spatial attention fields (AFs) to focus on suspicious local motion signals and integrate regional colliding information. 
	Using a nonlinear integration approach, we simulate the response of LPLC2 neurons and continuously extract local motion information from regions of interests. 
	This integration is more energy-efficient and effective than independently calculating the response of each LPLC2 neuron.
	
	To evaluate the proposed mLPLC2 model, we used computer-generated stimuli with various motion features to test its functional characteristics. 
	We then assessed its performance in dynamic natural scenarios with multiple looming objects, as well as in real-world UAV scenes facing diverse collision threats. 
	The mLPLC2 model showed notable improvements over traditional LGMD-based systems, particularly in its flexibility to focus on fast-approaching objects emanating from any region of the visual field.
	
	The structure of this paper is as follows: Section~\ref{Sec: related work} reviews typical bio-inspired models for looming detection. 
	Section~\ref{Sec:Proposed Model} describes the proposed model in detail. 
	Section~\ref{Sec:Experiments} presents experimental evaluations of the model. 
	Finally, Section~\ref{Sec:Dicussion} summarizes the findings and discusses future directions.
	
	\section{Related Work}
	\label{Sec: related work}
	
	This section provides a brief review on the most relevant works regarding (1) development of LGMDs for detecting looming objects, and (2) functional characteristics of individual LPLC2 neurons and their population dynamics.
	
	\subsection{Bio-inspired LGMD Looming Detector}
	\label{subSec:LGMD, ...}
	
	The looming-sensitive LGMD1-based neural network, inspired by the locust visual system, was introduced by Rind and Bramwell in 1996 \cite{1996_Rind_LGMD}. 
	They demonstrated that visual information is processed through excitatory inputs and lateral inhibition within the presynaptic structure, which shapes LGMD1's unique selectivity. 
	This numerical model responds most strongly to looming objects, exhibits brief excitation to receding objects, and generates weak, short-term responses to translating stimuli. 
	LGMD2, a neighboring counterpart to LGMD1, also functions as a looming detector. 
	Its presynaptic excitation is thought to be mediated by ON and OFF cells \cite{1976_on_off}. 
	In 2015, Fu and Yue developed the first LGMD2-based visual neural network, modeling the ON and OFF mechanisms \cite{2015_on_off_Fu}. 
	This model separates luminance changes into parallel channels, encoding excitatory and inhibitory signals through spatiotemporal computations similar to the LGMD1 model. 
	It successfully reproduces LGMD2's specific selectivity, responding exclusively to light-to-dark luminance changes, such as those caused by dark objects with centrifugal expanding edges \cite{Fu-ON/OFF-2023}.
	
	However, both LGMD1 and LGMD2 computational models demonstrate limitation on well distinguishing radial motion from other patterns, such as translation, compromising their selectivity for collision detection.
	
	\subsection{LPLC2 Physiology}
	\label{subSec:Ultra-selective, ditrectional, LPLC2, LPLC2 population}
	
\begin{figure}[t]
		\centering
		\includegraphics[width=0.5\textwidth,keepaspectratio]{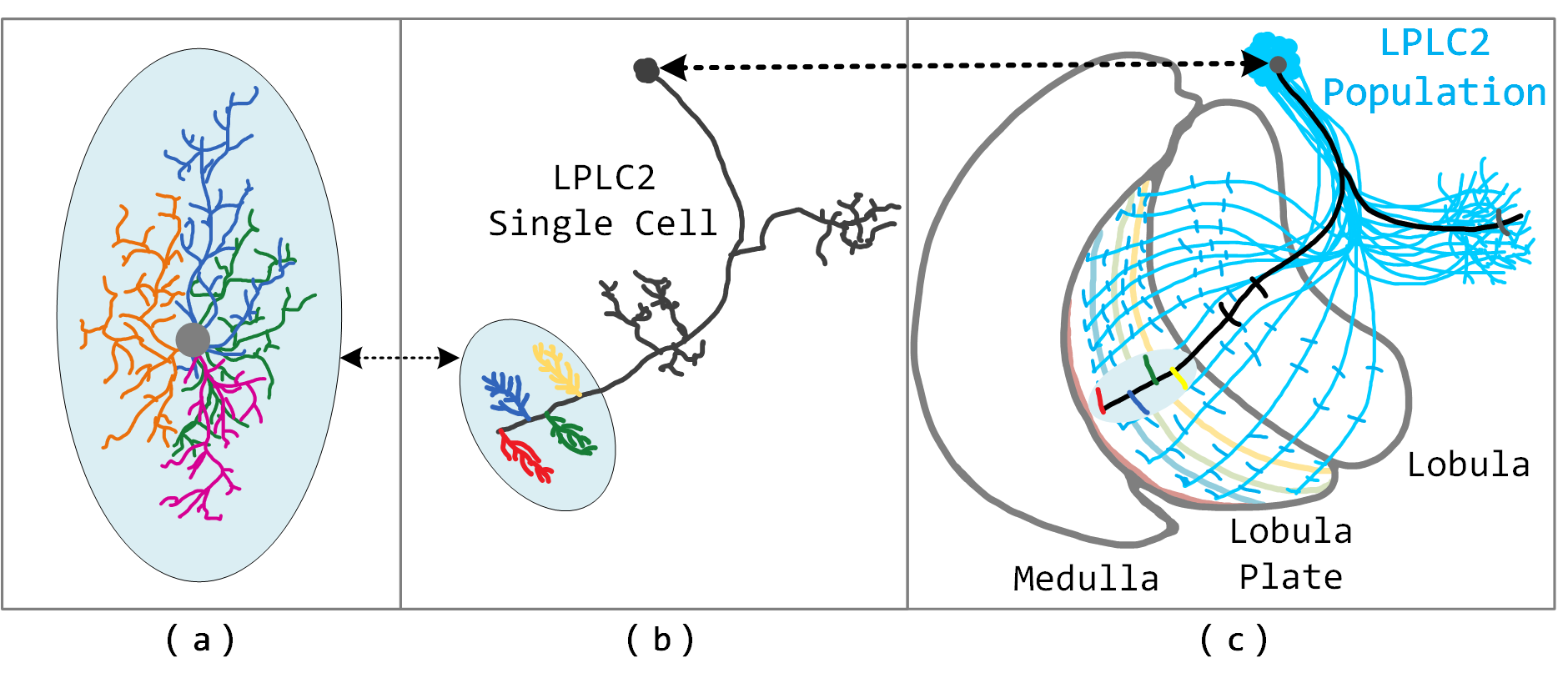}
		\caption{
			The schematic anatomical structure of LPLC2 neurons is as follows: 
			(a) Each arm of LPLC2's cross-shaped primary dendrites ramifies in one of the lobula plate layers and extends along that layer's preferred motion direction. 
			(b) A single LPLC2 cell integrates directional signals from four sub-layers of lobula plate. 
			(c) LPLC2 cells are a population of $\sim$80 visual projection neurons, with their dendrites collectively covering the lobula plate.
		}
		\vspace{-10pt}
		\label{Figure_LPLC2_Physiology}
	\end{figure}

	In 2017, Klapoetke et al. identified an ultra-selective ensemble of looming-detecting neurons, LPLC2, in \emph{Drosophila} \cite{2017_LPLC2_klapoetke}. 
	Individual LPLC2 neurons possess dendritic arbors that form four distinct fields, each occupying a separate layer of the lobula plate. 
	Within each lobula plate layer, the directionally selective signals from T4 and T5 neurons are gathered. 
	More specifically, these presynaptic, directionally selective neurons are tuned to one of the cardinal directions: front-to-back, back-to-front, upwards, and downwards, serves as the local optic flow sensors, directly providing inputs to LPLC2 ensemble \cite{2013_T4/T5_ON/OFF,2023_how_flies_see_motion} (Fig. \ref{Figure_LPLC2_Physiology}a).
	
	The four dendritic branches of an LPLC2 neuron form a radial, outward-oriented structure that aligns with the edges of a looming object expanding from the center of its receptive field (Fig. \ref{Figure_LPLC2_Physiology}b). 
	The direction of maximum outward dendritic spread is aligned with the preferred direction of T4 and T5 cells terminating in specific layers \cite{2017_LPLC2_klapoetke}. 
	This dendritic architecture is highly specialized for detecting centrifugal motion from a small looming object whose visual expansion matches the receptive field center of the LPLC2 neuron.
	
	By contrast, common motion stimuli, such as translation, only partially align with this structure and are suppressed by strong inhibitory responses to centripetal motion \cite{2017_LPLC2_klapoetke}. 
	In 2022, Hua et al. further refined the understanding of LPLC2's ultra-selective characteristics via highly nonlinear neural computation, demonstrating its strong responsiveness to dark, centroid-emanated centrifugal motion patterns while remaining nearly unresponsive to motion originating outside its receptive field center \cite{2022_LPLC2_Hua_Mu}.
	
	However, a single LPLC2 neuron can only process visual information within its receptive field, a limited area approximately \(60^{\circ}\) in diameter \cite{2022_LCs_klapoetke}. 
	In contrast, LGMD neurons receive stimuli through a single giant dendritic arbor that spans the entire visual field of one compound eye \cite{1996_Rind_LGMD}. 
	While this broad coverage enables LGMD neurons to detect looming objects, it limits their ability to localize or distinguish multiple objects in different parts of the visual field. 
	Accurately and continuously identifying multiple colliding objects from various locations across the visual field remains a significant challenge to existing bio-inspired computational models.
	
	The LPLC2 population, consisting of approximately $\sim$80 visual projection neurons, offers a potential solution. 
	With their dendrites collectively covering the lobula plate, the population enables regional detection of motion across the entire visual field \cite{2017_LPLC2_klapoetke,2023_how_flies_see_motion}. 
	This study explores how LPLC2 neurons distributed across different regions respond to motion using bottom-up salience generation, thus provides insights into their potential to separate global visual information into regional processing.
	
	\begin{figure*}[t]
		\centering
		\vspace{-20pt}
		\includegraphics[width=\textwidth,keepaspectratio]{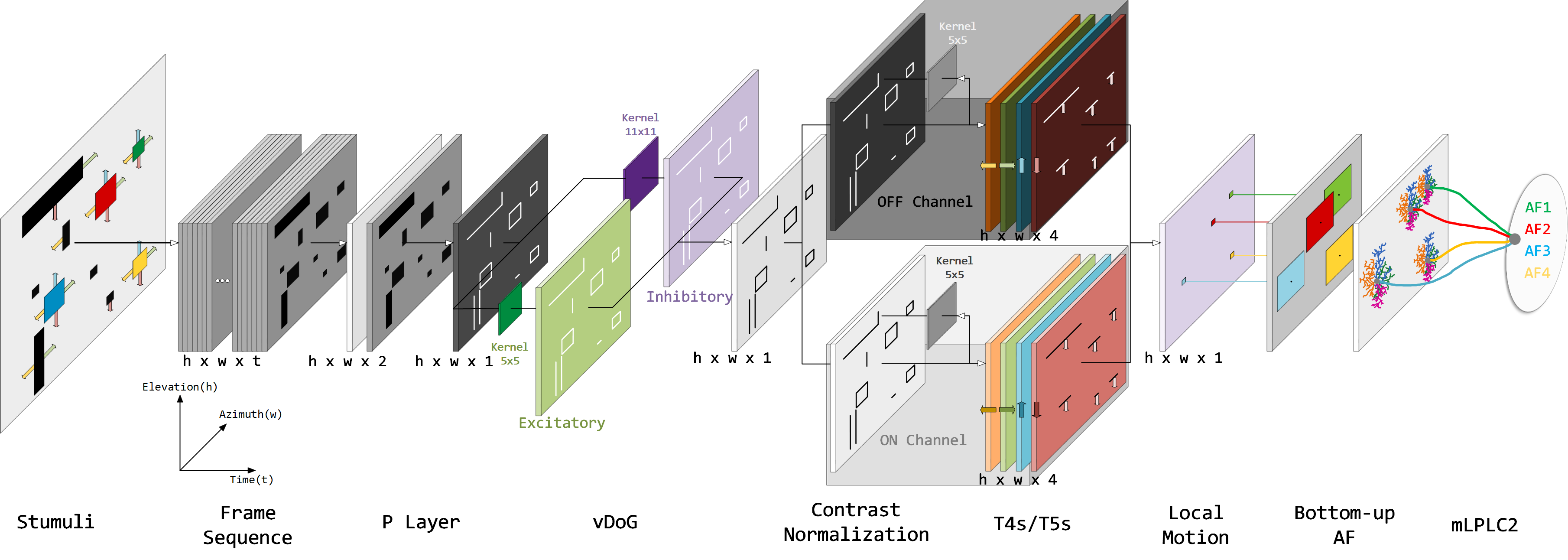}
		\caption{
			The flowchart of the proposed mLPLC2 neural network outlines the following steps. 
			The stimulation is input into the neural network, where the difference between consecutive frames is calculated to extract edge expansion information. 
			Excitation and inhibition signals from the lamina monopolar cells (LMCs) are integrated using the vDoG (variant Difference of Gaussian) simulation with polarity selectivity. 
			Motion signals representing brightness increases and decreases are sent into the ON and OFF channels, respectively, where they undergo contrast normalization. 
			A triple-correlation model simulates T4 and T5 cells to extract motion in four cardinal directions. 
			Directional signals are integrated to generate the local motion signal. 
			The most suspicious local motion signal is selected to drive a bottom-up process, identifying a new spatial attention field (AF). 
			The mLPLC2 model non-linearly integrates motion signals from multiple AFs, enabling the detection and localization of multiple looming objects.
		}
		\vspace{-10pt}
		\label{Figure_Model}
	\end{figure*}
	
	\section{Formulation of the Neural Network}
	\label{Sec:Proposed Model}
	
	This section begins with an introduction to the computational presynaptic neuropils of LPLC2. 
	Following this, we describe the multiple-attention mechanism employed in the model. 
	The algorithm for the proposed mLPLC2 model is outlined in pseudocode in Algorithm~\ref{Algo_LPLC2_Multi-Attention}, with the parameter settings detailed in Table~\ref{Table_Params}.
	
	\subsection{Modeling of Presynaptic Neuropils}
	\label{subSec_Pre-synaptic}
	
	The retinal layer is derived from the compound eye structure of \emph{Drosophila}, arranged as a 2D-matrix of photoreceptors($\mathit{P}$), with the amount corresponding to the resolution of the input visual images. 
	Each photoreceptor captures the gray-scale brightness and computes the difference with the previous frame to extract the brightness changes, which is defined as
	\begin{equation}
		\label{Eq_P_Layer}
		L(x,y,t) = P(x,y,t) - P(x,y,t-1),
	\end{equation}
	where $\mathit{P(x,y,t)}$ and $\mathit{P(x,y,t-1)}$ represent the gray-scale brightness of two successive frames.
	
	In the lamina layer, a spatial mechanism named variant of Difference of Gaussians (vDoG) with ON/OFF-contrast selectivity \cite{2020_Drosophila_motion_direction} is used to simulate excitation and inhibition signal produced by large monocular cells (LMCs) \cite{1991_monopolar_cells}. 
	The vDoG removes redundant background motion and enhances the selectivity to moving edges. 
	The excitation can be calculated as 
	\begin{equation}
		\label{Eq_vDoG_convolution}
		E(x,y,t)=\!\iint_{D_e}\,\! L(x+u,y+v,t) \cdot K_e(u+r_{D_{e}},v+r_{D_{e}}) \,\mathrm{d}u\,\mathrm{d}v,
	\end{equation}
	where $r_{D_{e}}$ stands for the radius of excitatory zone around LMC located at $(x,y)$. 
	The inhibition $I(x,y,t)$ conforms to the same computation that is omitted here. 
		The Gaussian kernel $K_e$ can be calculated as 
	\begin{equation}
		\label{Eq_vDoG_kernel}
		K_{e}(u,v)=\frac{1}{2\pi {\sigma_{e}}^2}\;e^{-\frac{u^2+\,v^2}{2{\sigma_{e}}^2} },
	\end{equation}
	where $\sigma_e$ represents the standard deviation of the inner Gaussian. 
	The outer inhibitory Gaussian kernel $K_i$ can be calculated in the same way.
	Subsequently in the vDoG algorithm, the broader inhibitory Gaussian is subtracted from the narrower excitatory one to enhance the extraction of edge expansion feature $\big(L_{vDoG}(x,y,t)\big)$, and the operation is consistent with the previous research
	 \cite{2020_Drosophila_motion_direction}.	
	After that, the LMCs split signals into ON and OFF pathways, an essential phenomenon observed in biological visual systems \cite{2010_ON/OFF_Drosophila,2020_fly_ON/OFF,2023_how_flies_see_motion}. 
	The ON pathway carries information about luminance increments, whereas the OFF pathway signals luminance decrements. This mechanism can be realized by `half-wave' rectifier which is defined as
	\begin{equation}
		\label{Eq_ON/OFF}
		\begin{alignedat}{2}
			&L_{ON}(x,y,t)=\frac{L_{vDoG}(x,y,t) + \left|L_{vDoG}(x,y,t)\right|}{2}, \\ 
			&L_{OFF}(x,y,t)=\frac{\big|\left|L_{vDoG}(x,y,t)\right| - L_{vDoG}(x,y,t)\big|}{2}.
		\end{alignedat}
	\end{equation}
	
	In the medulla layer, trans-medullary, medulla-intrinsic, centrifugal, and complex tangential cells, together, transmit and process temporally filtered signals \cite{2023_how_flies_see_motion}. 
	Within this process, contrast normalization is applied to both ON and OFF pathways through spatial integration of feedback neural signals \cite{2021_contrast_vision,Fu-Array}. 
	Let $N$ stands for both ON and OFF interneurons, the entire process is defined as
	\begin{equation}
		\label{Eq_contrast_tanh}
		N(x,y,t)=tanh\Bigg( \frac{N(x,y,t)}{\varepsilon + \hat{N}(x,y,t)} \Bigg),
	\end{equation}
	\begin{equation}
		\label{Eq_contrast_conv}
		\hat{N}(x,y,t)=\!\iint_{D_c}\!N(x+u,y+v,t) \cdot K_c(u+r_{D_{c}},v+r_{D_{c}}) \,\mathrm{d}u\,\mathrm{d}v,
	\end{equation}
	where $D_c$ stands for the impact zone around a medulla cell located at $(x,y)$ with radius at  $r_{D_{c}}$. 
	The Gaussian kernel $K_c$ can be calculated similarly to Eq.~\ref{Eq_vDoG_kernel}.
	
	Subsequently, the visual signal is transmitted to the direction-selective T4 cells in the medulla and T5 cells in the lobula \cite{2013_T4/T5_ON/OFF,2023_how_flies_see_motion}. 
	We use a triple-correlation model \cite{Clark2014(fly-human-estimation)} based on Hassenstein-Reichardt Correlators (HRC) \cite{1956_Emd_HR} to simulate every single type of T4(T5) cell perceives motion from one out of four cardinal directions in ON(OFF) channel \cite{1989_EMD_principles,2011_EMD_ON/OFF}. 
	Each T4/T5 samples from two locations: the position itself, and a certain distance in one of the four cardinal directions. 
	Here we apply a first-order low-pass filter to obtain the delayed signal $D(x,y,t)$ as 
	\begin{equation}
		\label{Eq_EMD_lp_delay}
		\begin{aligned}
			D(x,y,t) = \alpha_1 \cdot N(&x,y,t) + (1 - \alpha_1) \cdot N(x,y,t-1), \\
			&\alpha_1 = \tau_i / (\tau_i + \tau_1),
		\end{aligned}
	\end{equation}
	where $\tau_1$ is a delay and $\tau_i$ is the time interval between two consecutive frames in milliseconds.
	
	Taking the rightward motion detector $M_{r_i}(x,y,t)$ as an example here, it correlates triple input signals from two positions, at different time as
	\begin{equation}
		\label{Eq_EMD_HRC}
		\begin{aligned}
			M_{r_i}(x,y,t)&=N(x,y,t)\cdot D(x,y,t)\cdot D(x+sd,y,t) \\
			-\beta\cdot N&(x+sd,y,t)\cdot D(x,y,t)\cdot D(x+sd,y,t), 
		\end{aligned}
	\end{equation}
	where $\beta$ is a bias coefficient at null-preferred motion. 
	$i$ stands for current step in the dynamic delay mechanism and $N(x,y,t)$ denotes the contrast-normalized information from ON/OFF channels. 
	For each local T4(T5) cell, we sum the outputs of HRCs at various distances from the cell itself as the local motion perception result for T4(T5) cells. 
	Next, the outputs of a local cell correlating neighboring cells at various distances are accumulated, and a first-order low-pass filter is then applied to introduce a delay, yielding the directional signals for T4 (T5) cells, as the following:
	\begin{equation}
		\label{Eq_EMD_T4/T5_delay}
		\begin{aligned}
			&\hat{M}(x,y,t)=\sum_{i=1}^{N_c}M_{r_i}(x,y,t)\;,\quad\alpha_2 = \tau_i / (\tau_i + \tau_{N_c}),\\
			&T4_r(x,y,t)=\alpha_2 \hat{M}(x,y,t) + (1 - \alpha_2) \hat{M}(x,y,t-1),\\
		\end{aligned}
	\end{equation}
	where $N_c$ is the number of connected neurons and $\tau_{N_c}$ is the corresponding dynamic delay for T4(T5) cells in milliseconds.
	
	In four distinct layers of the lobula plate, LPLC2 neuron receive inputs from four sub-populations of both T4 and T5 cells \cite{2017_LPLC2_klapoetke,2022_LPLC2_Hua_Mu}. 
	The local directional motion signals from T4 and T5 cells at each position $(x, y)$ at time $t$ are integrated into $LM_\mathcal{V}(x, y, t)$ $\big ($ $\mathcal{V}\in {\{r, l, d, u}\}$ represents rightward, leftward, downward and upward local motion, respectively$\big )$ to facilitate subsequent calculations. 
	Here, we take the rightward motion computation as the example:
	\begin{equation} 
		\label{Eq_Local_Motion_4_direction}
		\begin{aligned}
			&r(x,y, t)={T4_r(x,y,t)}^{\gamma_1 } + {T5_r(x,y,t)}^{\gamma_2 }, \\
			&l\,(x,y, t)={T4_l\,(x,y,t)}^{\gamma_1 } + {T5_l\,(x,y,t)}^{\gamma_2 }, \\
			&\;\; LM_r(x,y, t)\!=\! f\big( r(x,y, t)-l(x,y, t) \big),
		\end{aligned}
	\end{equation}
	where $f(\cdot)$ denotes the Leaky-ReLU activation function. 
	$r(x, y, t)$ and $l(x, y, t)$ represent the rightward and leftward motion information of T4/T5 cells integrated from the ON and OFF channels, respectively. 
	$\gamma_1$ and $\gamma_2$ are exponential weights for the ON and OFF channels.
	
	Subsequently, two directional systems integrate the magnitude of local motion $LM(x,y,t)$ at time t via horizontal motion $H(x,y,t)$ and vertical motion $V(x,y,t)$ as 
	\begin{equation}
		\label{Eq_Local_Motion_Magnitude}
		\begin{aligned}
			&H(x,y, t)=max\big(LM_r(x,y, t),LM_l(x,y, t)\big), \\
			&V(x,y, t)=max\big(LM_d(x,y, t),LM_u(x,y, t)\big), \\
			&\;\; LM(x,y, t)=||H^2(x,y, t) + V^2(x,y, t)||_2. 
		\end{aligned}
	\end{equation}
	
	\subsection{Bottom-up Multi-Attention Based Nonlinear Integration}
	\label{subSec_Attention_Mechanism}
	
	Next, we introduce the multiple-attention mechanism driven by bottom-up motion salience that equips the LPLC2 model with the capability for multi-target looming detection, forming the mLPLC2 framework. 
	When multiple objects approach from various regions of the visual field, the proposed mechanism enables mLPLC2 to promptly detect and focus on these objects from the very first moment.
	
	At each time t, we search for suspicious motion cues from the whole visual field. 
	When a target is on a direct approaching course, its edges will appear to expand outward from a single point in the observer's view. 
	The expanding edges recorded in the $LM$ 2D-matrix start from a single point and gradually increase in size. 
	We select $(\hat{x}, \hat{y})$ at time t that maximizes motion magnitude $LM(x,y,\text{t})$ as the centroid of new attention field, which is retinotopically aligned with the initial position of the expanding motion. 
	Then the new attention field $AF^{n + 1}(\hat{x},\hat{y})$, with its center$(\hat{x},\hat{y})$ and radius $R_{AF}$, for suspicious local motion $LM(\hat{x},\hat{y},\text{t})$ will be incorporated into the set of all existing attention fields $AF^{N}=\{AF^0(x_0,y_0), \cdots , AF^n(x_n,y_n)\}$, when the following conditions are satisfied:
	(1) the location does not belong to any existing AFs; (2) it has the highest local motion magnitude outside the existing AFs, and this magnitude exceeds the threshold $T_{a}$. 
	Accordingly, the whole process is defined as the following:
	\begin{equation}
		\label{Eq_new_AF}
		\begin{aligned}
			&\qquad \qquad \quad (\hat{x},\hat{y}) = arg\underset{x,y}{max}\big(LM(x,y,\text{t})\big), \\ 
			&\qquad (x,y)\in \complement_\Omega \big(AF^0(x_0,y_0)\cup\!\cdots\!\cup AF^n(x_n,y_n)\big), \\
			&AF^{N+1} \!=\! AF^N \cup AF^{n + 1}\!(x^{n+1},y^{n+1}), \text{if}\; LM(\hat{x},\hat{y},\text{t}) \!>\! T_{a}. \\
		\end{aligned}
	\end{equation}
	$\Omega$ denotes for the entire visual field and $AF^0(x_0,y_0)\sim AF^n(x_n,y_n)$ represent current existing attention fields.
	
	Next, directional signals covered by each existing attention fields $AF^N$ is integrated in a highly nonlinear manner. 
	Specifically for each LPLC2 cell corresponding to a single $AF^n$, each arm of its cross-shaped primary dendrites ramifies in one of the lobula plate layers and extends along that layer's preferred motion direction \cite{2017_LPLC2_klapoetke}. 
	Within $AF^n$, the motions toward four cardinal directions at time $t$ are mapped into four quadrant:
	\begin{equation}
		\label{Eq_Integration_4_Quadrants}
		\begin{aligned} 
			&Q_1^n(t) = \!\iint_{\Gamma_{1}^n}\,\! \big(LM_r(u,v,t) + LM_u(u,v,t)\big)\,\mathrm{d}u\,\mathrm{d}v; \\
			&Q_2^n(t) = \!\iint_{\Gamma_{2}^n}\,\! \big(LM_l(u,v,t) + LM_u(u,v,t)\big)\,\mathrm{d}u\,\mathrm{d}v; \\
			&Q_3^n(t) = \!\iint_{\Gamma_{3}^n}\,\! \big(LM_l(u,v,t) + LM_d(u,v,t)\big)\,\mathrm{d}u\,\mathrm{d}v; \\
			&Q_4^n(t) = \!\iint_{\Gamma_{4}^n}\,\! \big(LM_r(u,v,t) + LM_d(u,v,t)\big)\,\mathrm{d}u\,\mathrm{d}v,
		\end{aligned}
	\end{equation}
	where $\Gamma_{1}^n \!\sim\! \Gamma_{4}^n$ denotes the four quadrants in the $AF^n$ divided by the centroid of $AF^n$ located at $(\hat{x},\hat{y})$.
	
	Only when all four dendritic arbors have received their preferred motion signal from $AF^n$, its postsynaptic $LPLC2^n(t)$ will be activated. 
	The attention-driven regional information is integrated as
	\begin{equation}
		\label{Eq_Integration_LPLC2_3}
		\begin{aligned}
			LPLC2^n(t)&=bool\Big( {\textstyle \prod_{i=1}^{4} Q_i^n(t) } \Big) \\
			 		  &\times \big(Q_1^n(t)+Q_2^n(t)+Q_3^n(t)+Q_4^n(t)\big),
		\end{aligned}
	\end{equation}
	where $bool(\cdot)$ function outputs $1$ once none of $Q_1 \!\sim\! Q_4$ equals $0$. 
	Importantly, if the LPLC2 cell in current $AF^k$ remains inactive or shows minimal activation after a certain period, the $AF^k$ will disappear, mathematically expressed as follows:
	\begin{equation}
		\label{Eq_deField}
		AF^N \!=\! AF^N \setminus AF^k_{(x_k,y_k)}\;, \;\text{if}\;\;\Sigma^t_{i=t-d} LPLC2^k(i) < T_{d},
	\end{equation}
	where $T_{d}$ is a threshold and $d$ denotes the time duration over which responses are accumulated within the current AF. 
	However, the mLPLC2 model ensures that at least one AF is always retained.
	
	In summary, the proposed model continuously monitors potential collision areas, generating response information that provides detailed and precise data about looming objects in the given scenario. 
	The parameters of the mLPLC2 neural network are empirically derived, inspired by biological functions. 
	With its feed-forward structure, this model operates free of data, without the need of training.
	
	\begin{table}[!t]
		\vspace{-10pt}
		\caption{Parameter Configuration}
		\centering
		\begin{tabular}{l l l}
			\hline
			Parameter & Description & Value\\ 
			\hline
			$\{r_{D_{e}},r_{D_{i}},r_{D_{c}}\}$ & the impact radius of LMC/medulla cell &  \{5,11,5\}  \\
			$\{\sigma_e,\sigma_i,\sigma_c\}$ & standard deviation in $K_e,K_i,K_c$ &  \{10,20,20\}  \\
			$\varepsilon$ & coefficient in contrast normalizasion & 0.2 \\
			$N_c$ & the number of connected neurons & 5 \\
			$\{\tau_1,\tau_{N_c}$\} & time delay in milliseconds & \{80,80~40\} \\
			$\beta$ & bias coefficient for HRC & 1.5 \\
			$\{\gamma_1,\gamma_2\}$ & exponent on the ON and OFF signals & \{0.9,0.5\} \\
			$R_{AF}$ & the radius of a single attention field & 40 \\
			$T_{a}$ & threshold for establishing attention field & 10 \\
			$T_{d}$ & threshold for conserving attention field & 5000 \\
			$d$ & time length for accumulating response  & 10$\cdot \tau_i$ \\
			\bottomrule
		\end{tabular}
		\vspace{-10pt}
		\label{Table_Params}
	\end{table}
	
	\begin{algorithm}[!h]
		\caption{LPLC2 with Multi-Attention Mechanism}
		\label{Algo_LPLC2_Multi-Attention}
		\DontPrintSemicolon
		\For{t = 2 : T}
		{
			\KwInput{ $P(x,y,t)$ }
			\tcp{\textbf{Computation of Pre-synaptic Neuropils}}
			\tcp{Retina layer}
			
			
			Compute $L(x,y,t)$ via Eq.~(\ref{Eq_P_Layer});

			\tcp{Lamina layer}
			
			Compute $L_{vDoG}(x,y,t)$ via Eq.~(\ref{Eq_vDoG_convolution})$\sim $(\ref{Eq_vDoG_kernel});
			
			Split the visual signal into ON/OFF via Eq.~(\ref{Eq_ON/OFF});

			\tcp{Medulla layer}
			
			Compute HRCs via Eq.~(\ref{Eq_contrast_tanh})$\sim $(\ref{Eq_EMD_HRC});
			
			Compute directional-selective neurons via Eq.~(\ref{Eq_EMD_T4/T5_delay});		
			
			\tcp{Lobula plate layer}
			
			
			Integrate local motion via Eq.~(\ref{Eq_Local_Motion_4_direction}), (\ref{Eq_Local_Motion_Magnitude});

			\tcp{\textbf{Operation of Attention Fields}}
			
			\tcp{search for suspicious motion cue}
			$\textbf{bool} \;\; \text{NewAttention = 1;}\; $
			$double \; \text{Maxpoint = 0;} $
			
			\For{y = 1 : Height of visual field}
			{
				\For{x = 1 : Width of visual field}
				{
					\tcp{Judge whether to set a new AF}
					\If{(x,y) $\in$ any existing AF}
					{
						NewAttention = 0;
					}
					
					\tcp{localize the new \text{AF} centroid}
					\If{$MaxPoint^n(\hat{x},\hat{y},\textnormal{t}) < LM(x,y,\textnormal{t})$}
					{
						$MaxPoint^n(\hat{x},\hat{y},\textnormal{t}) = LM(x,y,\textnormal{t})$;
					}
					
				}\textbf{end for}
			}\textbf{end for}
			
			\tcp{Update AFs}
			\If{$\text{NewAttention}$}
			{
				Generate a new $AF^n$ at $MaxPoint^n(\hat{x},\hat{y})$;
				
			}

			\tcp{Integrate response for all existing AFs}
			\For{$AF^n$ = 1 : n}
			{
				Integrate response of $\text{AF}^n$ as $LPLC2^n(t)$ 
				
				\If{$\Sigma^t_{i=t-d} LPLC2^n(i) < T_{d}$}
				{Abandon $AF^n$}
				
			}\textbf{end for}
			
			\KwOutput{$LPLC2_{(x_1 ,y_1)}^1(t)\sim LPLC2_{(x_n ,y_n)}^n(t)$}
			
		}\textbf{end for}
		
	\end{algorithm}
	
	\section{Experimental Evaluation}
	\label{Sec:Experiments}
	
	In this section, we first explore the specific selectivity of mLPLC2 by setting up comparative experiments in Subsection~\ref{Experiment_basic}. 
	Then, we challenge mLPLC2 using multiple appearing objects in dynamic natural scenes in Subsection~\ref{Experiment_hyper}. 
	Finally, in Subsection~\ref{Experiment_UAV}, we evaluate the practicality of mLPLC2 using real-world footage captured from UAVs.
	
	\subsection{Model Characteristic Study}
	\label{Experiment_basic}
	
\begin{figure}[t]
		\centering
		\vspace{-20pt}
		\includegraphics[width=0.5\textwidth,keepaspectratio]{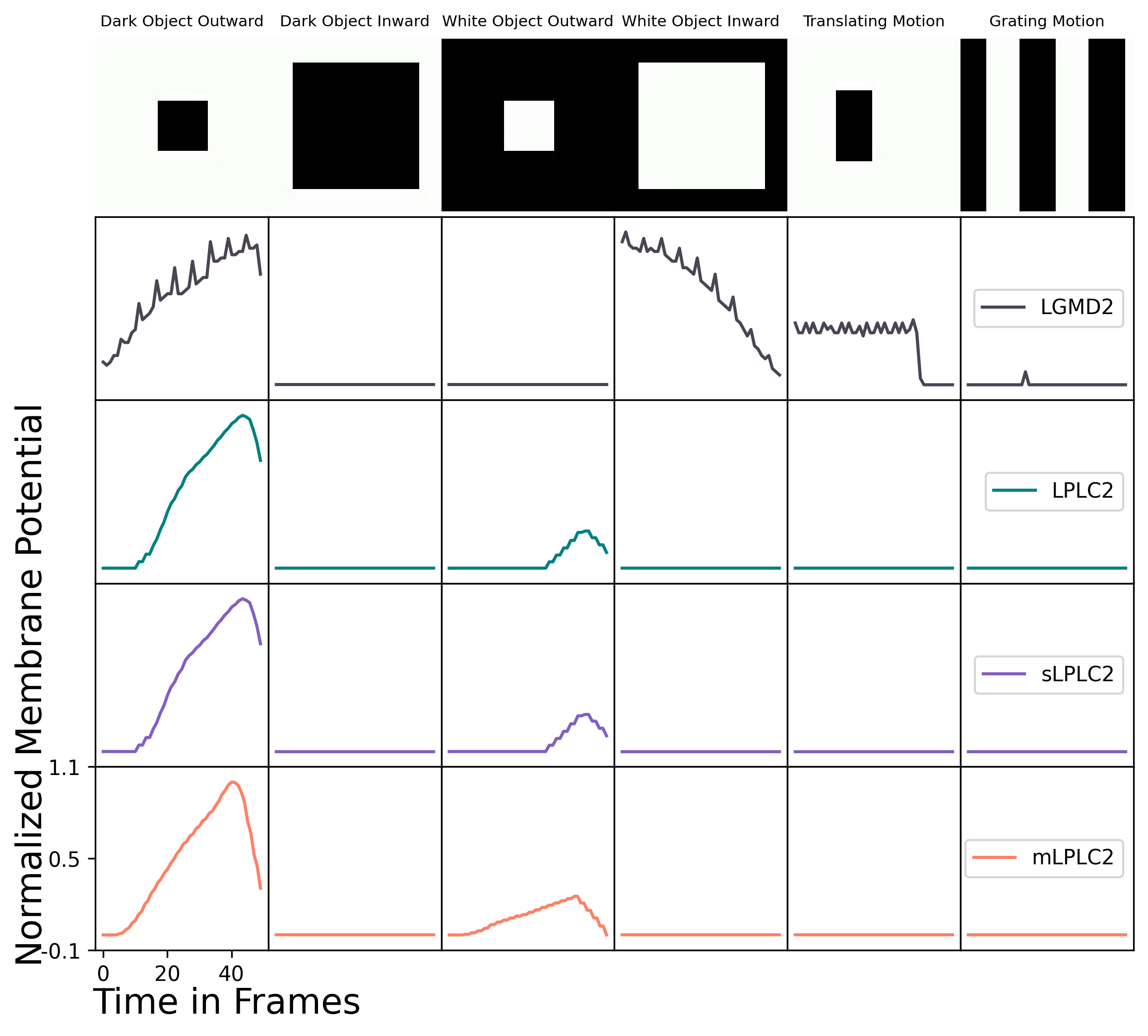}
		\caption{
			Different models' responses to various motion characteristics 
			(sLPLC2 is a constrained version of mLPLC2, limited to at most a single AF). 
			In the figure, each square represents a complete experiment, with the x-axis indicating time in frames and the y-axis representing the corresponding membrane potential of the models. 
			The LPLC2-based neural network model exhibits selectivity to only approaching stimuli.
			}
		\label{Figure_basic}
		\vspace{-10pt}
	\end{figure}

	To investigate the basic functionality of mLPLC2, we tested its responses to dark objects approaching, receding, bright objects approaching, receding, translating motion, and large-scale grating motion (input sampling resolution at 100$\times$100, length at 50 frames, and frequency at 33 Hz). 
	The LGMD2 model \cite{2015_on_off_Fu} with specific selectivity and the LPLC2 model \cite{2022_LPLC2_Hua_Mu} without the attention mechanism were used as comparisons. 
	Notably, we also limit the AF of mLPLC2 to a maximum of $1$, allowing single-attention LPLC2 (sLPLC2, with its single AF center generated at the centroid of the global local motion signal values) to participate in the experiment.
	The experimental results are presented in Figure~\ref{Figure_basic}, with snapshots of different motion patterns displayed in the first row.
	
	The experimental results clearly demonstrate the similar selectivity of LPLC2, sLPLC2, and mLPLC2. 
	Specifically, they all exhibit the strongest response to black objects approaching from the center of the visual field and show weaker responses to bright approaching objects. 
	They are not responsive to receding objects with a contracting trend, nor do they react to translating or large-scale grating motion. 
	In contrast, the LGMD2 model exhibited a markedly different response, reacting to bright, retreating objects. 
	Additionally, it was disturbed by motion patterns with translational characteristics, such as translating motion. 
	Evidently, the LPLC2-based model demonstrates superior performance in detecting only approaching objects.
	As the edge of an object exceeds the receptive field of a single LPLC2 cell, the cell can no longer detect the edge expansion, causing the membrane potential to decrease. 
	This phenomenon is evident from frame 40 onward in columns 1 and 3 of Figure~\ref{Figure_basic}.
	Overall, the performance of sLPLC2 and mLPLC2 is very similar with the LPLC2 model, showing an extremely selective response to expanding objects.
	The black object that elicited the strongest response was selected to conduct subsequent experiments. 
	
	\subsection{Performance Against Shifting Natural Scenes}
	\label{Experiment_hyper}
	
\begin{figure}[t]
		\centering
		\vspace{-20pt}
		\includegraphics[width=0.5\textwidth,keepaspectratio]{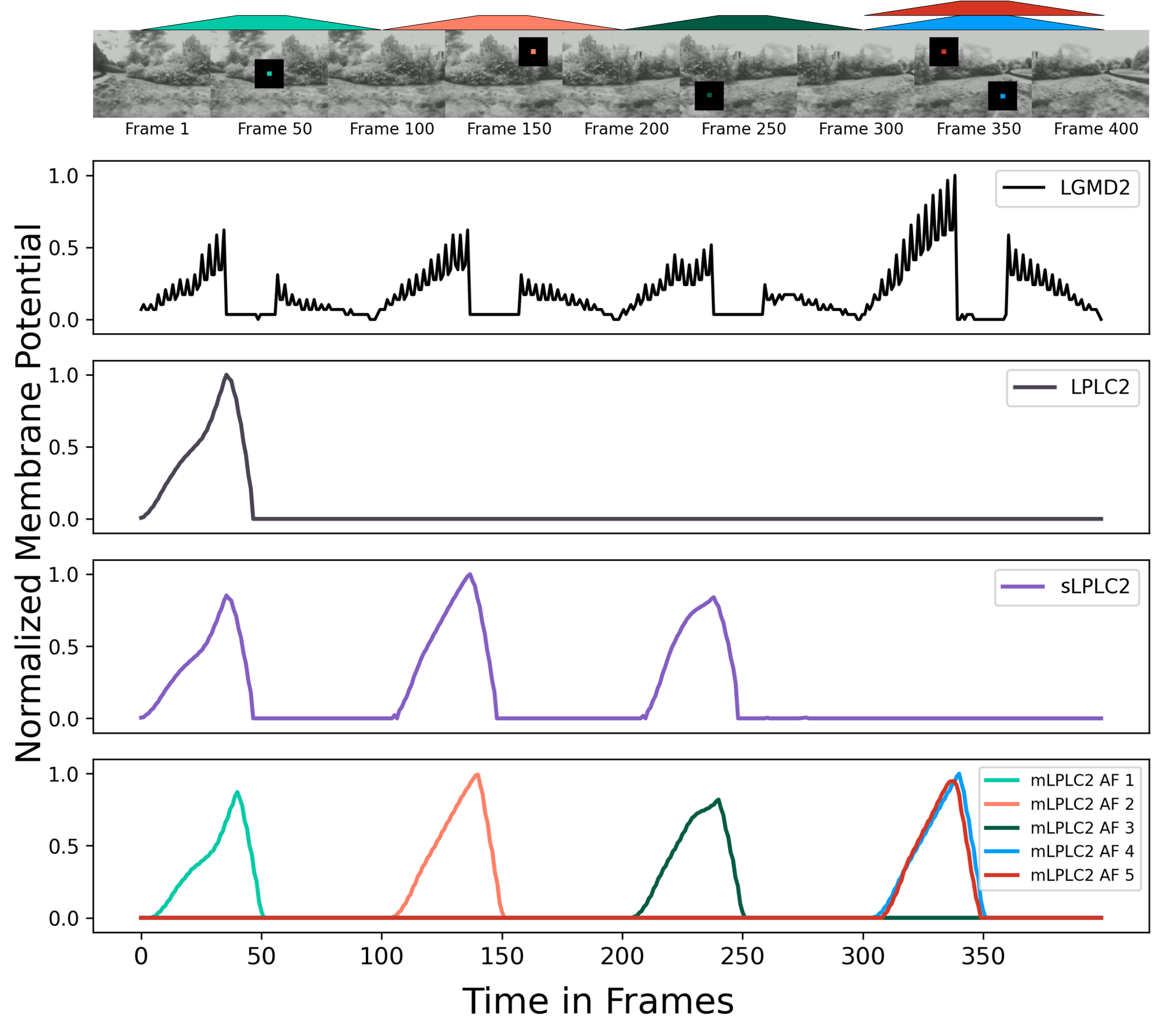}
		\caption{Different models' responses to objects from various regions of the visual field across four phases. 
		The horizontal and vertical axes represent time in frames and membrane potential, respectively. 
		The widths and heights of the differently colored shapes above the graph indicate the appearance period and size of the respective objects. 
		Each center of expansion is marked with corresponding color. 
		In the face of multi-target motion, only mLPLC2 can accurately distinguish and identify the looming objects.}
		\label{Figure_4_period}
		\vspace{-10pt}
	\end{figure}

	Hua et al.'s LPLC2 model only responds to expanding objects located at the center of the visual field \cite{2022_LPLC2_Hua_Mu}. 
	In contrast, we explore mLPLC2's response to approaching objects with expanding centers located at various positions. 
	Stimuli consisting of four phases were created to simulate the motion of objects confined to different regions of the visual field (input sampling resolution at 320$\times$240, length at 400 frames, and frequency at 33 Hz).
	Each phase is characterized by specific moving objects: 
	(1) a dark object located at the center of the visual field; 
	(2) a dark object in the first quadrant; 
	(3) a dark object in the third quadrant; and 
	(4) two dark objects located in the second and fourth quadrants, respectively. 
	Each phase contains 100 frames, with the first 40 frames showing the objects approaching, frames 40 to 60 featuring the objects stationary, and the final 40 frames depicting the objects receding. 
	The background features a real-world scene shifting left at a constant speed. 
	Notably, in the fourth phase, the movements of the two objects occur synchronously.
	The results are presented in Figure~\ref{Figure_4_period} with key frames.
	
	Based on the comparative experiments, it was observed that under cluttered background interference, the LGMD2 model exhibited a slight response to retreating dark objects.
	LGMD2 can respond to dark objects in any region of the visual field, although its membrane potential fluctuations are more pronounced compared to the LPLC2-inspired model. 
	However, LGMD2 cannot effectively distinguish the multi-object looming motion in the fourth stage.
	The ultra-selective LPLC2-inspired model remain silent to the receding inward motion of the black objects in the later stages of each phase. 
	However, LPLC2 model can only fire against the centroid-emanated centrifugal motion, which is insufficient to address the looming objects in stages 2, 3, and 4. 
	The sLPLC2 reacts timely and accurately to outward motion emanated from different visual regions in stages 2 and 3.
	However, sLPLC2 shows no response to motion in stage 4.
	This is because it locates the mass center of the global motion signal values as its AF center, which means that the two expanding rectangles cannot stimulate all four arbors of LPLC2, resulting the silence. 
	In contrast, the mLPLC2 model with multiple-attention mechanism accurately responds to objects emanating from different locations, including two objects approaching simultaneously across various phases. 
	Throughout the entire experimental process, mLPLC2 correctly generated 5 distinct AFs localizing the different looming objects over time.
	
\begin{figure}[t]
		\centering
		\vspace{-20pt}
		\includegraphics[width=0.5\textwidth,keepaspectratio]{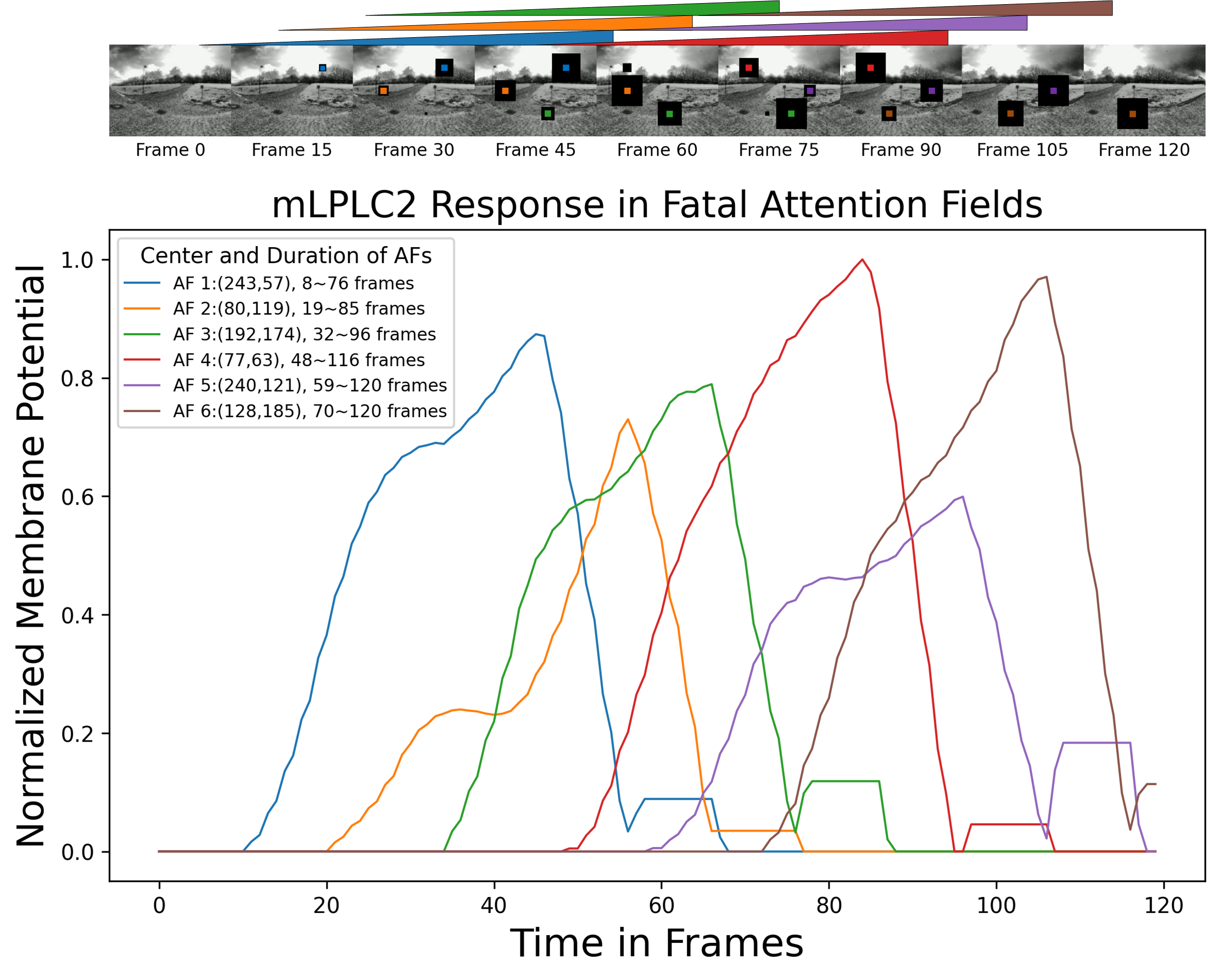}
		\caption{mLPLC2's responses to objects originating from different regions of the visual field and at various times, amidst a shifting natural background. The presence period and size of each object are represented by the widths and heights of differently colored shapes at the top of the graph. The center positions and presence periods of AFs are shown in the legend, corresponding closely to the 6 looming objects. In response to these 6 objects, mLPLC2 continuously detects and locks onto multiple looming targets, demonstrating effective responsiveness.}
		\label{Figure_6_objects}
	\end{figure}

	To further investigate the multi-target detection capability of mLPLC2, we generated more complex simulated stimuli. 
	Six dark objects appeared at frames 5, 15, 25, 45, 55, and 65 in different positions within the global visual field of mLPLC2 then move centrifugally. 
	The background featured a chaotic real outdoor scene shifting to the left at a constant speed. 
	The experimental results are presented in Figure~\ref{Figure_6_objects} with key frames and the duration of object presence displayed above.
	
	When encountering collision objects that exhibit a centripetal outward expansion pattern at various positions and times, mLPLC2 can lock onto these objects within a maximum of 7 frames (23 ms) to establish an AF for each of looming targets. 
	Interestingly, this AF persists for approximately 20 frames (66 ms) after the object disappears. 
	This persistence is attributed to the dynamic background at the object's location, which induces a luminance change in the Retina layer upon the object's vanishing, ultimately leading to the excitation of the corresponding AF. 
	This observation is illustrated in the post-peak regions of the curves shown in Figure~\ref{Figure_6_objects}.
	Due to the influence of high-contrast scenes, the edges of certain collision objects may merge with the background during their expansion, and the edges of multiple simultaneously appearing objects may overlap at certain moments. 
	This complexity results in varying peak responses across different Attention Fields. However, thanks to the ultra-selectivity of LPLC2 neurons and the well-defined attention rules (as outlined in Eq.~\ref{Eq_deField}), mLPLC2 remains capable of recognizing these challenging collisions and establishing fatal AFs.
	
	Experimental evidence demonstrates that, in complex dynamic real-world scenes, mLPLC2 can accurately localize and respond to every single looming object.
	
	\subsection{Real World Challenges in UAV Scenarios}
	\label{Experiment_UAV}
	
\begin{figure}[t]
		\centering
		\vspace{-20pt}
		\includegraphics[width=0.5\textwidth,keepaspectratio]{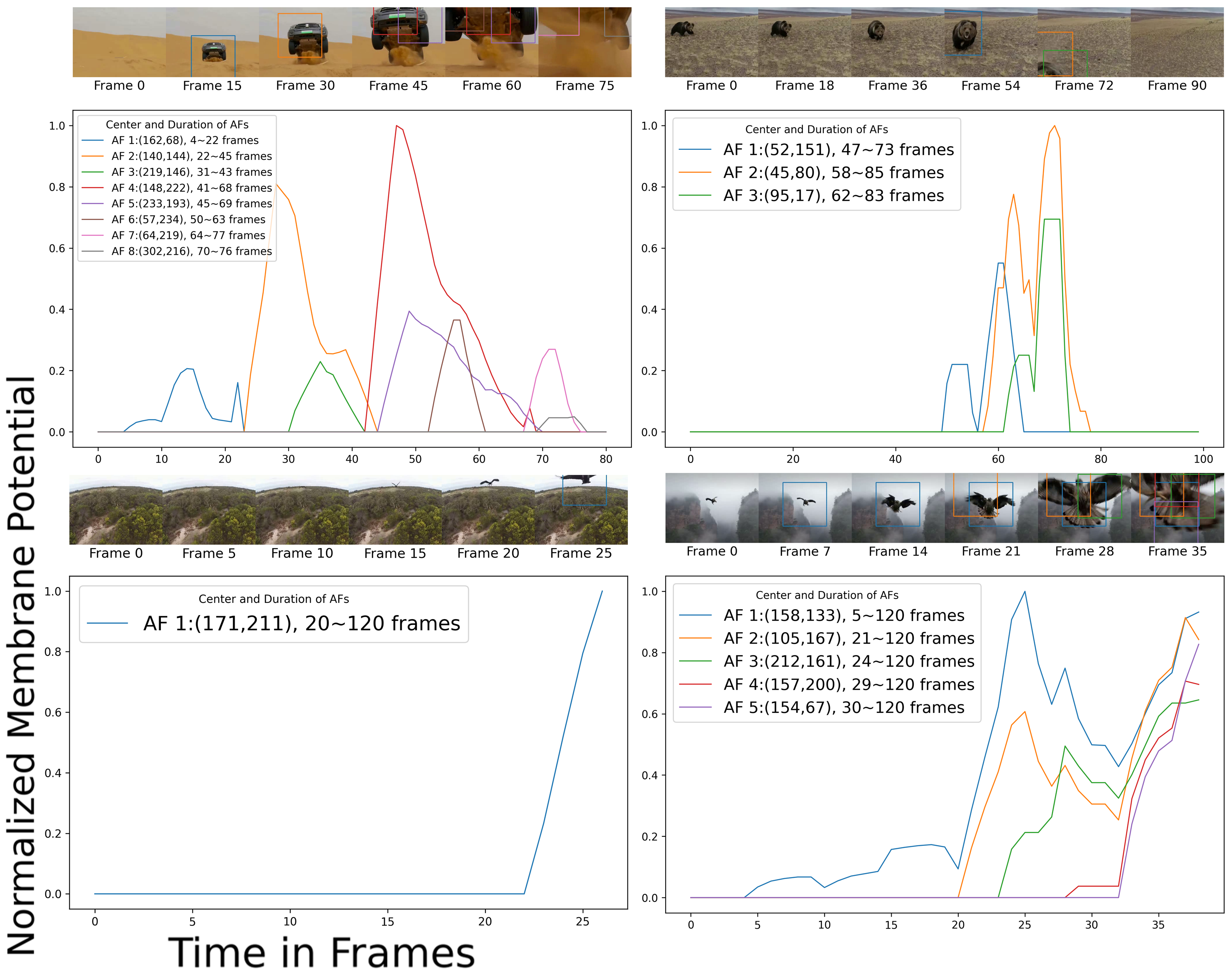}
		\caption{
			This figure illustrates mLPLC2's responses to dangerous colliding scenes captured by UAVs. 
			In each subplot, the x-axis represents time in frames, and the y-axis denotes the normalized membrane potential. 
			The first subplot depicts a drone filming a leaping car. 
			The second subplot shows a drone facing an approaching bear.
			The third and fourth subplots feature drones confronted by raptors.
			This figure demonstrates that mLPLC2 effectively detects and localizes real-world hazards encountered by UAVs.}
		\label{Figure_UAV}
	\end{figure}

	Finally, we evaluated the practical performance of mLPLC2 using $4$ common UAV footage examples. These videos include various scenarios where drones encounter hazards, such as approaching predator attacks (Figure~\ref{Figure_UAV}).
	
	mLPLC2 demonstrated timely responses to various real-world threats faced by drones.
	However, two notable observations emerged from the experiments: 
	First, mLPLC2 generates multiple AFs when facing large objects in its field of view. 
	This occurs because the expanding edges of the approaching object surpass the AF of an individual LPLC2 cell’s receptive field. 
	In such cases, a single AF is insufficient to capture the full expansion motion, prompting the formation of additional AFs along the expanding edges. 
	This effect is evident in the first subplot with the leaping car and in the fourth subplot with the approaching raptor.
	
	Second, mLPLC2 also generates multiple AFs when the expansion center of a moving object shifts. 
	This behavior results from the unique motion integration pattern of LPLC2 cells, which only process motion expanding radially from the AF center in four primary directions. 
	When the expansion center shifts, the original AF can no longer track the motion effectively, leading to the formation of new AFs. 
	This phenomenon is observed in the first and second subplots, where the expansion centers of both the upward-leaping car shift and the running bear, triggering multiple AFs in response.
	
	\section{Conclusion and Discussions}
	\label{Sec:Dicussion}
	
	This paper introduces an attention-driven mLPLC2 neural ensemble model for continuously detecting and localizing multiple looming targets. 
	Based on the experimental results, the following conclusions can be drawn: 	
	(1) The mLPLC2 neural network retains the ultra-selectivity of the LPLC2 model for radial motion expansion. 
	(2) With its multiple-attention mechanism, mLPLC2 accurately and promptly responds to multiple radial motion objects originating from different focused regions across the visual field. 
	
	The model's ultra-selectivity for expanding objects makes it exceptionally well-suited for collision detection tasks. 
	The attention mechanism addresses a key limitation of the original single-LPLC2 model, which was restricted to responding only to objects emanating from the center of the whole visual field. 
	This advancement enables precise, large-scale collision detection and localization across multiple regions, potentially providing vehicles or robots equipped with this system with more detailed and comprehensive collision scenario information. 
	As a result, it facilitates more intelligent and precise avoidance maneuvers during navigation. 
	
	The mLPLC2 model's capabilities make it particularly advantageous for UAV applications in three-dimensional motion spaces, where wide visual coverage and precise multi-dimensional collision avoidance actions are essential. 
	However, the mLPLC2 model inherits certain limitations from the biological LPLC2 neurons in \emph{Drosophila}. 
	Since individual LPLC2 neurons are confined to processing motion information within their receptive fields, they cannot track objects moving beyond these local regions. 
	Furthermore, existing nonlinear integration methods heavily rely on the biological structure of LPLC2's dendrite. 
	To address these challenges, future work would explore combining the global motion cue detection of the LGMD model with the mLPLC2 framework. 
	This integration sheds light upon enhancing the system's ability to handle moving and expanding objects across the entire visual field in complex and dynamic scenes.

	\section*{Acknowledgment}
	This research has been supported by the National Natural Science Foundation of China under Grant No. 62376063. 
	\textcolor{blue}{\textit{Corresponding author: Qinbing Fu} (\url{qifu@gzhu.edu.cn})}
%
%
%
%
%

	\bibliographystyle{IEEEtran}
	\bibliography{reference.bib}
	
\end{document}